\documentclass[conference]{IEEEtran}
\IEEEoverridecommandlockouts
\usepackage{cite}
\usepackage{amsmath,amssymb,amsfonts}
\usepackage{algorithmic}
\usepackage{graphicx}
\usepackage{textcomp}
\usepackage{booktabs}
\usepackage{textcomp}
\usepackage{makecell}
\usepackage{comment}
\usepackage{multirow}

\usepackage{xcolor}
\def\BibTeX{{\rm B\kern-.05em{\sc i\kern-.025em b}\kern-.08em
    T\kern-.1667em\lower.7ex\hbox{E}\kern-.125emX}}

\usepackage{fancyhdr}
\begin{document}

\title{Take it Personally: The Limits of General SSL Representations for Real-Life PPG Emotion Detection\\

\thanks{This work was partially supported by 
Polish National Science Centre, Poland, project no. 2024/53/B/HS6/01256;
the statutory funds of the Department of Artificial Intelligence, Wroclaw University of Science and Technology;
the Polish Ministry of Education and Science within the programme “International Projects Co-Funded”;
the European Union under the Horizon Europe, grant no. 101086321 (OMINO). However, the views and opinions expressed are those of the author(s) only and do not necessarily reflect those of the European Union or the European Research Executive Agency. Neither the European Union nor European Research Executive Agency can be held responsible for them.
}}

\author{\IEEEauthorblockN{Dominika Kunc}
\IEEEauthorblockA{\textit{Dept. of Artificial Intelligence} \\
\textit{Wrocław University of Science}\\
\textit{ and Technology}, Poland \\
dominika.kunc@pwr.edu.pl}

\and
\IEEEauthorblockN{Przemysław Kazienko}

\IEEEauthorblockA{\textit{Dept. of Artificial Intelligence} \\
\textit{Wrocław University of Science}\\
\textit{ and Technology}, Poland \\
przemyslaw.kazienko@pwr.edu.pl}

\and
\IEEEauthorblockN{Stanisław Saganowski}

\IEEEauthorblockA{\textit{Dept. of Artificial Intelligence} \\
\textit{Wrocław University of Science}\\
\textit{ and Technology}, Poland \\
stanislaw.saganowski@pwr.edu.pl}
}

\maketitle
\thispagestyle{fancy}

\begin{abstract}
   While Self-Supervised Learning (SSL) effectively extracts general representations from noisy, unconstrained physiological signals such as photoplethysmography (PPG), its suitability for highly subjective tasks remains unproven. In this work, we evaluate the efficacy of PPG-based SSL for real-life intense emotion detection. First, we pretrain a Real-Life PPG encoder (RL-PPG) on unconstrained, real-life data. As a rigorous sanity check, we demonstrate that these representations transfer exceptionally well to an objective physical activity recognition task, yielding almost 5-fold increase in performance over baselines in a leave-one-subject-out evaluation (LOSO). However, when applied to a~subjective real-life emotion detection task, these same general representations fail to surpass naive baselines under the LOSO protocol. Using an Across-Time validation strategy, we establish that incorporating an individual's personal data during fine-tuning is the main driver of predictive performance, outweighing the benefits of population-level pretraining. Ultimately, our findings indicate that in the evaluated scenario, general SSL representations may be insufficient for subjective affective inference, suggesting that personalization is likely a key component for real-world emotion recognition. To support future research, we share the code and pretrained RL-PPG~encoder~weights.

\end{abstract}
\section{Introduction}
Growing awareness of mental health's impact on overall well-being has highlighted the need for continuous monitoring tools. While consumer-grade wearables successfully provide detailed physical health feedback through physiological tracking, comparable tools for mental health support remain limited. Bridging this gap is crucial, as emotional awareness plays a vital role in stress regulation, psychological resilience, and well-being~\cite{salovey1990emotional,martins2010comprehensive,zeidner2012emotional}.

Affective computing seeks to address this by inferring affective states from physiological and behavioral signals. However, physiological responses to emotion are highly variable across individuals and contexts, supporting a constructionist view of emotion~\cite{hoemann2020context}. Many popular laboratory benchmarks fail to account for this variability by using the external stimulus or performed activity as the ground truth label. Consequently, models often learn to recognize the stimulus rather than the subject's actual subjective experience. Because individuals respond differently to the same stimulus based on their unique backgrounds, beliefs, and past experiences~\cite{kunc2024embracing}, these models struggle to generalize outside controlled settings.

To address these limitations, we utilize LarField -- a real-life affective dataset collected in unconstrained, everyday settings\cite{komoszynska2024designing}. Given that real-world affective experiences are highly personalized and cannot be mapped to universal categories, we simplify our problem formulation to a binary distinction: the presence or absence of an intense emotional experience. This approach reduces reliance on specific, subjective emotion labels while remaining relevant for mental health applications, where detecting emotional anomalies is valuable. Furthermore, it accounts for individual and cultural differences in emotional interpretation and intensity.

Because labeling unconstrained real-world data is expensive and difficult, self-supervised learning (SSL) has emerged to leverage the vast amounts of unlabeled data. While SSL is the standard for learning generalizable, transferable patterns in Computer Vision and Natural Language Processing, affect-related physiological signals present a unique challenge. Unlike a cat in an image or a verb in a sentence, the physiological patterns of affect are deeply subjective. Despite growing interest in SSL for physiological time series, it remains unclear whether these general representations can capture the subjectivity of everyday affective experiences.

This work makes three key contributions to real-world affective computing. First, we systematically evaluate the efficacy of real-life PPG SSL representations, demonstrating their robust transferability to objective physical tasks but notable limitations in subjective emotion detection. Second, through Across-Time validation, we provide evidence that fine-tuning with subject-specific data outperforms generalized pretraining, highlighting the potential necessity of personalized modeling. Finally, we open-source our pretrained RL-PPG encoder weights\footnote{Repository link: https://github.com/Emognition/RL-PPG}, while the LarField dataset is currently being prepared for public release.
 \vspace{-2mm} 

\section{Related Work}
\subsection{SSL for Physiological Time-Series}
SSL has become an effective and popular methodology for learning representations from physiological time series. In the domain of electrocardiogram (ECG) analysis, several SSL frameworks have already demonstrated strong generalization. For instance, CLOCS leverages temporal and spatial augmentations to learn representations that successfully transfer across various tasks \cite{kiyasseh2021clocs}. WildECG further extends this line of work by applying state-space modeling to large-scale, real-world ECG data \cite{avramidis2024scaling}. CM-TCC utilizes real-life ECG and accelerometer in a cross-modal learning, incorporating behavioral context into physiological representation \cite{kunc2026dominance}.

Despite these advances, SSL for wearable PPG signals remains less explored, likely due to their lower signal quality and greater susceptibility to noise compared to ECG signals. Consequently, existing approaches often include PPG as part of multimodal setups rather than treating it as a standalone modality \cite{dissanayake2022sigrep}. For example, the COCOA model learns cross-modal representations via contrastive learning and achieves strong classification performance with high label efficiency \cite{deldari2022cocoa}. Nevertheless, PPG-specific SSL has recently shown significant clinical promise. Models such as UPR-BP use standalone PPG representations for noninvasive blood pressure estimation, achieving medical-grade accuracy across diverse datasets \cite{chenbin2024uprbp}.

To address the challenges of modeling time series without relying on signal-specific assumptions, general-purpose SSL frameworks like TS-TCC \cite{eldele2021time} provide a robust, modality-agnostic alternative. By combining temporal and contextual contrasting, TS-TCC has demonstrated competitive performance across multiple datasets. It matches fully supervised models and shows particularly strong results in low-label and transfer learning settings, making it a highly suitable backbone for noisy, real-life PPG data collected via consumer wearables. 

Taken together, the success of these methods in handling signal noise, subject variability, and limited annotations strongly suggests that SSL should be an excellent tool for tackling our core problem: real-life emotion detection from wearable data.
\subsection{Emotion Recognition using Physiological Signals}
Emotion recognition from physiological signals is a cornerstone of affective computing. Early progress relied heavily on controlled laboratory datasets like DEAP~\cite{koelstra2011deap}, AMIGOS~\cite{miranda2018amigos}, Emognition~\cite{saganowski2022emognition}, and WESAD~\cite{schmidt2018introducing}, which supported diverse modeling approaches ranging from feature-based classifiers to deep learning~\cite{santamaria2018using,siddharth2019utilizing,harper2020bayesian}. However, laboratory studies are inherently constrained by researchers' \textit{a priori} assumptions. Explicitly designing stimuli to elicit predefined states limits the natural complexity of expressible experiences~\cite{posner1968genesis,barrett2022context}. Conversely, real-life settings lack experimental control, introducing profound variability rooted in individual circumstances, appraisals, and environments~\cite{kunc2024embracing,behnke2022ethical}.

The rise of consumer-grade wearables has naturally shifted research toward capturing these unconstrained everyday emotions~\cite{exler2016wearable,schmidt2019multi,schmidt2018labelling}. While these unobtrusive devices open new opportunities in healthcare and human-computer interaction~\cite{schmidt2019multi,dao2018healthyclassroom,exler2016wearable}, transitioning to real-world affect recognition introduces severe challenges: physiological signals suffer from motion noise, ground-truth labels are sparse and subjective, and affective experiences vary drastically across individuals~\cite{kunc2024embracing}. Moreover, daily-life affective experiences differ significantly in intensity and clarity from artificially induced in-lab states.

Consequently, laboratory-trained models rarely generalize to real-life scenarios, highlighting the critical need to explicitly account for inter-subject variability and inherent subjectivity~\cite{kunc2024embracing,behnke2022ethical}. This gap directly motivates our methodology: rather than relying on laboratory datasets or fine-grained categories that struggle to generalize in the wild, we utilize real-world data for both SSL pretraining and downstream classification, focusing on the binary detection of subjectively intense emotional events as a more realistic target for noisy physiological signals.

 \vspace{-2mm} 

\section{SSL Representation}
 \vspace{-1mm} 

\subsection{Pretraining Dataset}
The self-supervised pretraining was conducted using a subset of the LarField dataset collected in daily life \cite{komoszynska2024designing}. The dataset covers a four-week monitoring period and includes data from 48 participants. The subjects were using a custom-made mobile application \cite{kunc2023emognition, saganowski2022lessons}, which enabled collection of continuous physiological signals, i.e., PPG (25 Hz), ACC (50 Hz) from  Samsung Galaxy 3 smartwatches, and ECG and ACC data (130 Hz) from Polar H10 chest straps, along with affect-related momentary self assessments. After in-depth explanations (presentations, written materials), all the subjects signed an informed consent. The study was approved by the Ethical Committee. The participants were compensated  with the amount of up to 150 USD based on their engagement in study (more details can be found in Ethical Impact Statement). To support model development and evaluation, the data were partitioned at the subject level: unlabelled recordings from 22 individuals (12 female, mean age 29.4 $\pm$ 9.9, 2,246,171 10s PPG windows) were used for SSL pretraining, data from two participants (1 female, mean age 49.0 $\pm$ 0.0, 236,772 10s PPG windows) served as a validation set, and the remaining 24 subjects (10 female, 2 preferred not to say, mean age 25.8 $\pm$ 8.0) were reserved as a held-out test set. The splitting strategy aimed to balance age and gender distributions across subsets while prioritizing a high number of completed self-reports in the test group. Because real-world emotion labels are highly subjective, we intentionally allocated half of the available subjects to the test set as a smaller test set risks biasing results toward individual quirks.

The preprocessing steps for this pretraining dataset included window splitting, signal cleaning, and resampling. We decided to use non-overlapping 10-second windows, as this window length was commonly adopted in prior work on SSL for physiological time series \cite{sarkar2020self,vazquez2022transformer,avramidis2024scaling}. We applied a simple quality check, which excluded any samples with irregular lengths caused by sampling inconsistencies. Signal cleaning was performed using the NeuroKit2 toolkit \cite{makowski2021neurokit2}, followed by resampling to 25 Hz. Finally, each participant's data were normalized independently using z-score standardization. Preprocessing was intentionally kept lightweight to maintain the real characteristics of the signals, considering that real-world data often contains more noise compared to laboratory-grade signals.
    \vspace{-2mm} 

\subsection{Architecture}

\begin{figure}[htbp!]
  \vspace{-3mm} 

  \centering
  \includegraphics[width=0.7\columnwidth]{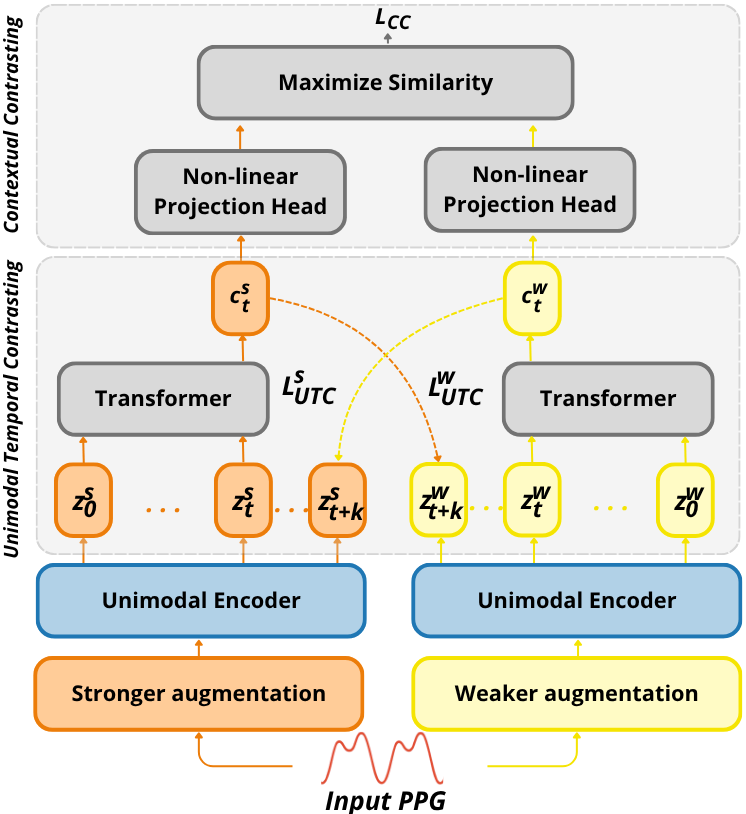}
  \caption{The RL-PPG architecture. Adapted from TS-TCC \cite{eldele2021time}.}  \label{fig:architecture}
  \vspace{-1mm} 

\end{figure}

We adopt the foundational principles of the TS-TCC framework \cite{eldele2021time}, a contrastive self-supervised learning method for time-series data, as the basis for our proposed RL-PPG model (presented in Fig. \ref{fig:architecture}). TS-TCC relies on two complementary contrastive objectives: temporal contrasting, which encourages the model to predict future representations from past context, and contextual contrasting, which pulls together augmented views of the same signal and pushes apart any other pair of signals. The TS-TCC framework has demonstrated strong performance across various time-series tasks, motivating our choice to adapt it rather than relying on a model explicitly tailored for clean, laboratory-grade data.

To tailor the encoder specifically for our large-scale, real-life dataset (comprising 10-second PPG windows sampled at 25 Hz), we implemented several structural modifications. We deepened the network by adding two convolutional layers and adjusted the kernel sizes and parameters to align precisely with our input shape, utilizing 10 timesteps in the temporal contrasting module. Furthermore, we replaced the original ReLU activations with GELU, which led to a more stable pretraining process. The resulting RL-PPG SSL encoder consists of five convolutional layers and approximately 345K trainable parameters, yielding a compact yet highly expressive representation model optimized for noisy, real-world physiology.

To generate contrasting views for the contextual contrasting module, we applied time-series augmentations, including jittering and permutation. Crucially, we adopted the lower-magnitude augmentation parameters. Keeping these parameters small is essential for our application, as it prevents augmentations from destroying the delicate morphological features of the PPG signal, which are vital for downstream physiological and affective inference. The pretraining phase was optimized using the Adam optimizer, with a learning rate of  $3\cdot10^{-3}$ for 40 epochs, which took around 14 hours on NVIDIA GeForce RTX 4090.

    \vspace{-1mm} 

\section{Experimental setup}

 Our experiments do not aim to prove the general efficacy of SSL~\cite{eldele2021time, kunc2026dominance}. Instead, we evaluate whether representations learned from unconstrained, noisy wearable data can successfully transfer to the deeply subjective task of real-life intense emotion detection. To avoid severe domain shifts and ensure an ecologically valid evaluation, we maintain strict domain consistency. We evaluate the model on downstream tasks using data collected under the same real-world conditions as the pretraining set, deliberately avoiding cross-domain transfer from clean, medical-grade laboratory data.
Crucially, before tackling the subjective problem of emotion classification, we validate our representations on an objective task: physical activity recognition. This serves as a rigorous sanity check, confirming that the pretrained encoder successfully extracts meaningful physiological patterns from real-life PPG signals before applying them to human affect.

    \vspace{-2mm} 

\subsection{Datasets and downstream tasks}

\subsubsection{ProSi}
The ProSi dataset \cite{polak2022processing} is a multimodal physiological dataset for physical activity recognition recorded in both laboratory and semi-naturalistic settings. It includes data from 11 participants. All of them were university students or employees associated with our research team. In our experiments, we consider only recordings obtained using the Samsung Galaxy Watch 3 smartwatch, which provides PPG measurements sampled at a rate of 25 Hz. This hardware setup is exactly the same as in the pretraining dataset, which guarantees fair evaluation. All signals were divided into 10-second-long non-overlapping windows, and they were preprocessed in the same way as the pretraining dataset.

The dataset consists of seven five-minute activities, grouped into \textit{stationary} and \textit{running} categories. The \textit{running} session included three treadmill-based activities performed at speeds of 1, 5, and 9 km/h. This tasks were designed to collect samples with increasing physical intensity and motion artifact levels. The \textit{stationary} protocol was conducted in a seated position and designed to induce varying levels of movement-related and physiological artifacts, including:
\textit{unconstrained sitting} - a baseline where participants remained still without any breathing instructions, \textit{breath} - paced breathing,  guided by a visual sinusoidal signal, \textit{arm raising} - periodic arm lifting to introduce changes in blood pressure, also guided by a visual sinusoidal signal, \textit{tap} - simulation of typical smartwatch interactions through repetitive finger tapping.

We evaluate this dataset on two downstream tasks: a binary classification task distinguishing \textit{stationary} from \textit{running} activities, and a multiclass classification task comprising seven individual activities.

\subsubsection{Real-life Intense Emotion}
\begin{figure}[htbp!]
  \vspace{-3mm} 

  \centering
  \includegraphics[width=0.83\columnwidth]{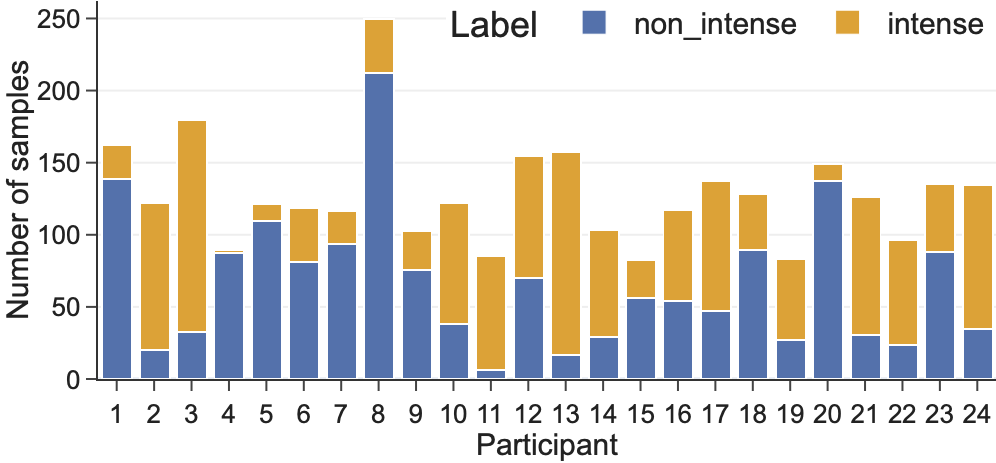}
    \vspace{-3mm} 

  \caption{Distributions of per-participant intense vs. non-intense emotion labels present in the Real-life Intense Emotion dataset.}  \label{fig:emo_labels_dist}
  \vspace{-3mm} 

\end{figure}

The Real-life Intense Emotion dataset is a hold-out test set from the pretraining LarField dataset. The self-assessment questionnaires were triggered by the mobile application based on predictions from a feature-based intense emotion detection ML model  \cite{kunc2023emognition}. The model continuously monitored the participants' physiology and triggered a questionnaire once it was confident of the presence of an intense emotional event. This mechanism enabled the collection of over 37\% more emotion-related self-assessments compared to a random trigger \cite{kunc2022real}. To ensure we included both intense and non-intense emotions, random triggering was also used. Every time the questionnaire was triggered, the subjects had to answer a brief question: "Did you feel an intense emotion $x$ minutes ago?", with possible answers "Yes", "No", and "I don't know" (discarded in this study). The $x$ was calculated based on time difference between current time when filling self assessments and the timestamp of the questionnaire trigger. In case of the random trigger it was simply just the randomly generated timestamp. For the ML model it was the middle timestamp of the prediction window which yielded high probability of intense emotion occurrence. Subjects could also report any event whenever they wanted. As this dataset is a hold-out test set, no data from these participants were present in the SSL pretraining, ensuring no information leakage. The data preprocessing was identical as in the pretraining dataset (10 second long, non-overlapping windows), and the split was performed in such a way to have the emotion timestamp exactly in the middle of the window.

 The dataset comprises 24 subjects and 3066 labeled samples (1475 intense and 1591 non-intense emotions). Overall, the dataset is well-balanced. However, the distribution of intense vs. non-intense emotion labels is highly variable across participants. The distribution of labels per subject is presented in Figure \ref{fig:emo_labels_dist}. Such an imbalance highlights the variability in people's real-life experiences, making emotion detection in real life a hard-to-generalize task.

\subsubsection{Real-life Physical Activity Recognition}
To generate orientation-invariant ground-truth labels for real-life binary physical activity, we developed an unsupervised heuristic pipeline using accelerometer data. For each temporal window from the Real-life Emotion Dataset, we computed the triaxial magnitude, $M = \sqrt{x^2 + y^2 + z^2}$, to mitigate device placement effects. We then calculated the variance of this magnitude as a proxy for movement intensity and applied a natural logarithmic transformation to address the inherent right skew. To account for inter-subject variability and ambiguous real-world motion, a three-component Gaussian Mixture Model (GMM) was fit to the log-transformed variances per subject. This GMM isolated three states: strictly stationary, an ambiguous intermediate state (e.g., sitting in a vehicle on a bumpy road), and strictly active. To maximize label purity, windows assigned to the intermediate component were discarded, resulting in a final dataset of 2165 samples (667 stationary and 1498 non-stationary). The remaining highly confident labels served as the objective task ground truth for evaluating our RL-PPG SSL representation.

    \vspace{-2mm} 

\subsection{Cross-Validation Scenarios}
    \vspace{-1mm} 

Careful splitting of data into training, validation, and test sets is particularly important when working with physiological time series. A simple random split at the window level can easily lead to information leakage, as neighboring samples from the same individual (e.g., within 1 minute) may end up in both the training and test sets. This can artificially inflate performance and give a misleading picture of a model's generalization. To avoid these issues and better reflect realistic evaluation settings, we use two validation strategies: LOSO and Across-Time (AT). 

\subsubsection{Leave-One-Subject-Out (LOSO)} In LOSO cross-validation, data from a single participant is held out for testing in each fold, ensuring that subjects are disjoint across the training, validation, and test splits. This rigorous setup simulates a real-life cold-start deployment, where no prior data for a new user is available. Although LOSO typically yields lower performance than conventional random splits, it provides a much more robust assessment of cross-subject generalization. We employ this strategy for both the physical activity and intense emotion detection tasks.

\subsubsection{Across-Time (AT)}

\begin{figure}[t]
\vspace{-6mm} 
  \centering
  \includegraphics[width=0.8\columnwidth]{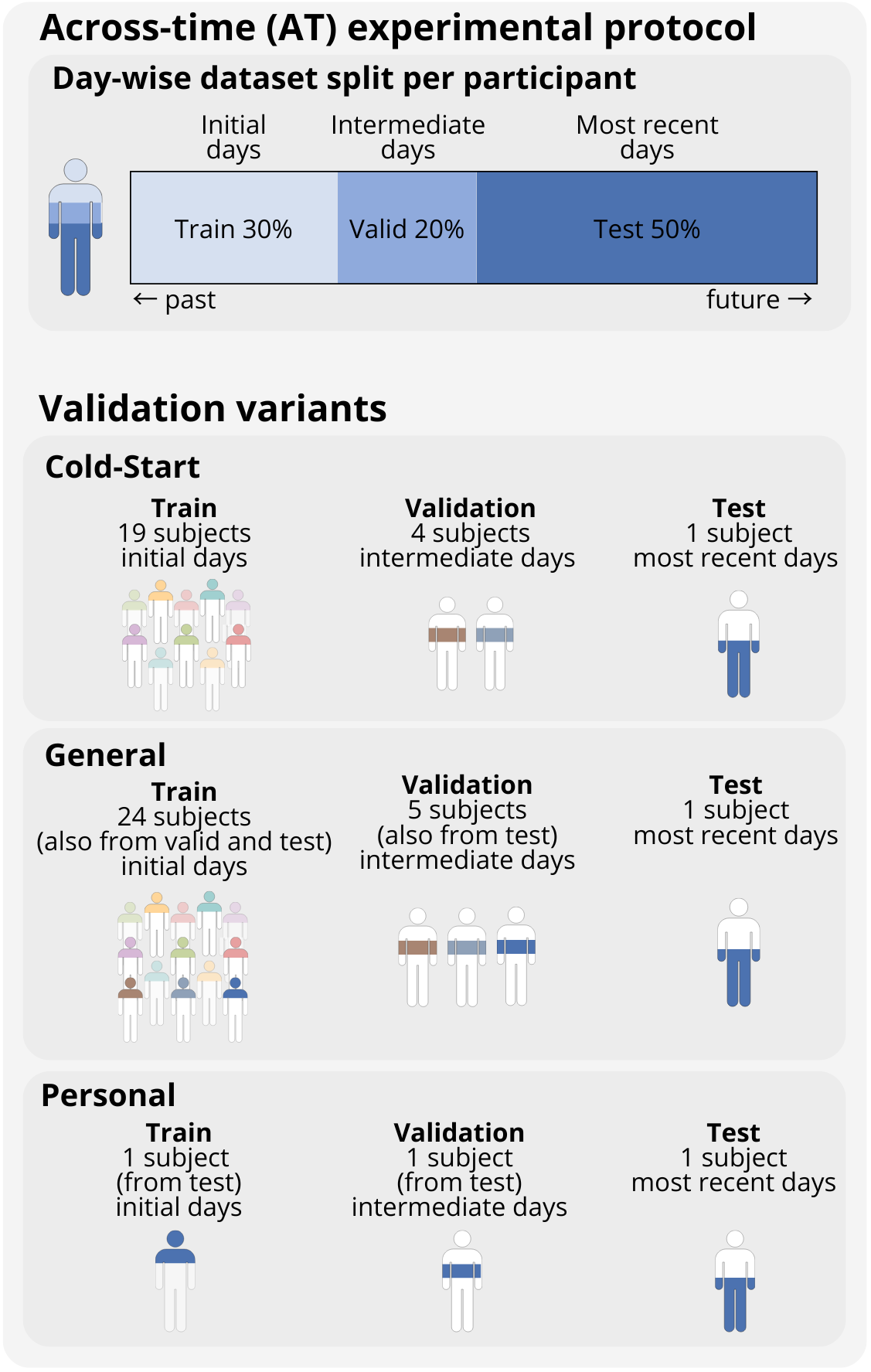}
    \vspace{-3mm} 
  \caption{Visual presentation of per-participant dataset split and folds in different validation variants in the Across-Time cross-validation (AT).}  \label{fig:across_time_explanation}
 \vspace{-5mm} 

\end{figure}

In the AT validation setup, we chronologically split each participant's data by full days. Because longitudinal studies typically suffer from declining self-report compliance over time, we adopt specific, non-traditional proportions to account for this: the initial 30\% of days form the training set, the subsequent 20\% the validation set, and the remaining 50\% the test set (Fig. \ref{fig:across_time_explanation}). The split proportions were intentionally designed to counterbalance declining compliance of self-reports, which is a known issue in longitudinal studies.
This temporal separation limits information leakage and allows us to evaluate model behavior under three scenarios of data availability:
\begin{itemize} 
\item \textit{Cold Start}: Excludes the test participant's data from training. This serves as a temporally constrained LOSO setup to evaluate pure generalization. 
\item \textit{General}: Combines the test participant's initial days with data from all other subjects. This reflects a realistic deployment scenario where prior user data is leveraged alongside a global dataset. 
\item \textit{Personal}: Trains exclusively on the test participant's initial data. This isolates participant-specific characteristics by eliminating inter-subject interference. \end{itemize}

Crucially, the test set for each participant remains identical across all three variants to ensure direct comparability. This AT procedure is conducted for all 24 participants in the real-life intense emotion dataset. 

    \vspace{-2mm} 

\subsection{Evaluation}
    \vspace{-2mm} 

For downstream classification, we employ a standard architecture consisting of a representation encoder followed by a multilayer perceptron (MLP) classification head. We evaluate two initialization strategies: training from scratch (FS Encoder) and initializing with pretrained weights (RL-PPG). As the architectures are \textit{exactly} the same, the only difference in the compared models is whether they are initialized with SSL pretrained weights or not. For both variants, we compare two optimization schemes: a \textit{frozen} setup, where only the MLP head is updated, and a \textit{fine-tuned} setup, where the entire network is optimized end-to-end.

To address the class imbalance, we train the downstream models using a weighted cross-entropy loss: \begin{equation} \mathcal{L}_{\text{WCE}} = - \frac{1}{N_{\text{total}}} \sum_{n=1}^{N_{\text{total}}} \sum_{i=1}^{N_{\text{classes}}} w_i \, y_{n,i} \log(\hat{y}_{n,i}), \end{equation} where $y_{n,i}$ and $\hat{y}_{n,i}$ are the ground-truth and predicted probabilities for sample $n$ and class $i$, respectively. The class-specific weights $w_i$ are inversely proportional to class frequencies: \begin{equation} w_i = \frac{N_{\text{total}}}{N_i \cdot N_{\text{classes}}}, \end{equation} where $N_{\text{total}}$, $N_i$, and $N_{\text{classes}}$ denote the total number of samples, the number of samples in class $i$, and the total number of classes. For hyperparameter tuning, we selected the optimal learning rate for each task based on average performance across a search grid of $\{0.1, 0.05, 0.005, 0.0005\}$.

To establish a quantitative lower bound, we contextualize model performance against two naive baselines: a majority-class predictor and a uniform random classifier. Across all experiments, we evaluate performance using the macro-averaged F1-score to account for class imbalance. To reflect cross-subject variability, all reported results represent the mean and standard deviation across the per-participant evaluation folds.
    \vspace{-2mm} 

\section{Results}

\subsection{Physical Activity Recognition Sanity Check}

\textbf{\textit{Laboratory Binary - stationary vs. running:}} Table~\ref{tab:loso_physical_activity} summarizes the results of our physical activity recognition sanity check. The findings clearly demonstrate the value of representation learning, particularly when initialized with SSL. In the frozen setting, both the FS and RL-PPG encoders substantially outperform the naive baselines. Notably, the frozen RL-PPG consistently surpasses its FS counterpart, yielding an F1-score improvement of roughly 0.08.


\
End-to-end finetuning yields further performance gains, adding +0.11 and +0.07 to the FS and RL-PPG encoders, respectively. Ultimately, the finetuned RL-PPG achieves an impressive macro F1-score of 0.90 on this binary task. This strong performance is remarkable because PPG does not directly capture kinematics. Instead, the model successfully learns to identify the indirect cardiovascular shifts (e.g., heart rate variations) associated with movement. This confirms that the pretrained RL-PPG encoder extracts highly meaningful physiological patterns.

\textbf{\textit{Laboratory Multiclass - seven distinct activities:}}
As expected, overall performance decreases in the more challenging multiclass setting. However, the underlying trends observed in the binary task remain perfectly consistent: SSL pretraining strictly outperforms training from scratch, and end-to-end finetuning yields the highest performance. 

The finetuned RL-PPG achieves a macro F1-score of 0.64, representing a nearly 4.6-fold increase over the uniform naive baseline (0.14). This substantial gain highlights the expressive power of the learned representations. To succeed here, the model must differentiate fine-grained physiological variations, such as specific breathing patterns, varying running paces, and subtle blood pressure changes, rather than just binary acceleration shifts.

\begin{table}[htpb]

\caption{F1 macro scores for the Physical Activity Recognition in Laboratory and Real-life, LOSO validation.  FS: from scratch.} 
\centering \begin{tabular}{lccc} 
\toprule 
& \multicolumn{2}{c}{\textbf{Laboratory}} & \textbf{Real-life} \\

\cmidrule(lr){2-3}\cmidrule(lr){4-4}

\textbf{Method} & \textbf{Binary} & \textbf{Multiclass} & \textbf{Binary} 
\\ 
\midrule 
\textbf{Naive Baseline} & & &\\
Majority-class & 0.36 $\pm$ 0.00 & 0.04 $\pm$ 0.00 & 0.41 $\pm$ 0.03 \\ 
Uniform & 0.50 $\pm$ 0.02 & 0.14 $\pm$ 0.01 & 0.49 $\pm$ 0.06  \\ \midrule \textbf{Frozen} & & & \\ 
FS Encoder & 0.75 $\pm$ 0.13 & 0.22 $\pm$ 0.07 & 0.81 $\pm$ 0.07  \\
RL-PPG & 0.83 $\pm$ 0.13 & 0.40 $\pm$ 0.13 & \textbf{0.88 $\pm$ 0.05} \\ \midrule
\textbf{Finetuned} & & & \\ 
FS Encoder & 0.86 $\pm$ 0.08 & 0.59 $\pm$ 0.11 & 0.85 $\pm$ 0.04 \\ 
RL-PPG & \textbf{0.90 $\pm$ 0.10} & \textbf{0.64 $\pm$ 0.10} & 0.87 $\pm$ 0.06 \\ \bottomrule \end{tabular} \label{tab:loso_physical_activity} 
\vspace{-3mm} 
\end{table}

\textbf{\textit{Real-life Binary - stationary vs. non-stationary:}}
Consistent with the laboratory findings, the models exhibit similar performance patterns on the real-life dataset, with SSL approaches consistently outperforming the FS variants in both frozen and finetuned configurations. The Frozen RL-PPG achieved the highest overall performance, reaching an F1-score of 0.88. While these results demonstrate robust real-world transferability, this strong performance may partially stem from the real-life dataset's binary labeling scheme, which simplifies the task by strictly isolating extreme stationary and non-stationary states. In contrast, the laboratory dataset includes more nuanced behaviors, such as altered seated activities, which complicate the models' decision boundaries.

Ultimately, this sanity check confirms that our RL-PPG model successfully extracts meaningful, highly discriminative patterns from noisy, real-world PPG data. This establishes a rigorous, scientifically sound foundation for applying the same methodology to our primary, inherently subjective target task: real-life intense emotion detection.

    \vspace{-1mm} 
\subsection{Intense Emotion Detection (LOSO)}
    \vspace{-1mm} 

The results for the subjective intense emotion detection task are presented in Table~\ref{tab:across_time}. In stark contrast to the physical activity sanity check, the models in LOSO validation setup fail to exhibit any meaningful performance gains. All evaluated methods, including naive baselines, stagnate within a narrow macro F1-score range of 0.43 to 0.48. The highest-performing configuration (the frozen FS encoder) outperforms the uniform baseline by a mere 0.02.

In the frozen setting, the RL-PPG fails to outperform the FS encoder, indicating that representations learned from other individuals' PPG data do not transfer effectively to this highly subjective task. Furthermore, when finetuned end-to-end, both the FS and RL-PPG encoders yield identical performance (0.46), exactly mirroring the uniform baseline. This suggests that under strict cross-subject evaluation, downstream training is insensitive to encoder initialization and instead of learning discriminative affective features, the network simply defaults to the underlying dataset distribution.

\begin{table}[htpb] 
\caption{F1 macro scores for Real-Life Intense Emotion Detection across, LOSO and Across Time validation. FS: from scratch.} 
\centering \resizebox{\columnwidth}{!}{
\begin{tabular}{lcccc} \toprule & \multirow{2}{*}{\textbf{LOSO}} &\multicolumn{3}{c}{\textbf{Across Time}} \\ \cmidrule(lr){3-5} \textbf{Method} & \textbf{}  & \textbf{Cold-Start} & \textbf{General} & \textbf{Personal} \\
\midrule 
\textbf{Naive Baseline} & & & \\ 
Majority-class &  0.43 $\pm$ 0.04 & 0.48 $\pm$ 0.16 & 0.48 $\pm$ 0.16 & 0.48 $\pm$ 0.16 \\ Uniform & 0.46 $\pm$ 0.05 & \textbf{0.49 $\pm$ 0.17} & \textbf{0.49 $\pm$ 0.17} & 0.49 $\pm$ 0.17 \\ 
\midrule 
\textbf{Frozen} & & & \\ FS Encoder & \textbf{0.48 $\pm$ 0.08}  & 0.42 $\pm$ 0.15 & 0.47 $\pm$ 0.12 & 0.53 $\pm$ 0.16 \\ RL-PPG & 0.45 $\pm$ 0.09  & 0.44 $\pm$ 0.14 & 0.47 $\pm$ 0.08 & 0.53 $\pm$ 0.16 \\ 
\midrule 
\textbf{Finetuned} & & & \\ FS Encoder & 0.46 $\pm$ 0.08 & 0.43 $\pm$ 0.09 & 0.44 $\pm$ 0.06 & \textbf{0.54 $\pm$ 0.16} \\ RL-PPG & 0.46 $\pm$ 0.08 & 0.44 $\pm$ 0.11 & 0.46 $\pm$ 0.08 & \textbf{0.54 $\pm$ 0.16} \\ 
\bottomrule 
\end{tabular} }
\label{tab:across_time} 
 \vspace{-3mm} 
\end{table}
 
Ultimately, these findings highlight a critical limitation: general representations learned from external populations may be insufficient for tasks rooted in subjective internal experiences. Because the LOSO protocol strictly excludes the test participant's data during training, the models lack the personalized context necessary to decode highly individualized emotional responses. Motivated by this severe drop in cross-subject generalization, we next introduce our Across-Time (AT) validation experiments to determine whether incorporating participant-specific data can bridge this subjectivity gap.

    \vspace{-1mm} 

\subsection{Intense Emotion Detection (Across-Time)}
    \vspace{-1mm} 

Table~\ref{tab:across_time} also summarizes the AT validation results for the real-life intense emotion detection task. Notably, the data volumes in this temporal setup are substantially lower than in the LOSO protocol: the training, validation, and test sets are approximately three, five, and two times smaller, respectively. Despite this substantial reduction in training data, the AT evaluation reveals crucial insights into how data availability impacts subjective task performance. 

\textbf{\textit{Cold-Start:}}
As this scenario strictly excludes the validation and test participants from the training set, it conceptually mirrors the LOSO setup, albeit with temporally constrained, smaller datasets. Consistent with the LOSO findings, none of the evaluated models surpasses the naive baselines. A simple uniform baseline achieves the maximum F1-score of 0.49. Finetuning the encoders provides negligible gains. This further reinforces our core observation: generalized representations aggregated from other individuals struggle to meaningfully transfer to highly subjective affective tasks.

\textbf{\textit{General:}}
In this variant, the training set includes the initial days of the test and validation participants, along with the general dataset. Compared to the Cold-Start scenario, we observe minor but consistent improvements, ranging from +0.01 (finetuned FS encoder) to +0.05 (frozen FS encoder). While this confirms that access to subject-specific data aids classification, the overall gains remain modest. Because the target participant's data constitutes only a small fraction of the aggregated training corpus, the dominance of external data likely introduces inter-subject noise, effectively weakening the personalized affective patterns.

\textbf{\textit{Personal:}}
In this setup, downstream training and validation rely exclusively on the test participant's own historical data, completely eliminating inter-subject noise. Here, we observe the most substantial performance gains, with finetuned encoders exceeding the uniform baseline by 0.05. Crucially, this is the only Across-Time scenario where the models consistently outperform naive guessing. When comparing the Personal setup directly to the Cold-Start baseline, all encoder variants show striking improvements ranging from 0.09 to 0.11. Notably, the FS and RL-PPG encoders yield identical performance in both the frozen and finetuned settings. This underlines that pretraining on external populations provides no supplementary benefit for this task. Instead, the performance gain is driven entirely by the isolation and utilization of highly personalized affective data. 
  \vspace{-3mm} 
\begin{figure}[htbp!]
  \centering
  \includegraphics[width=0.82\columnwidth]{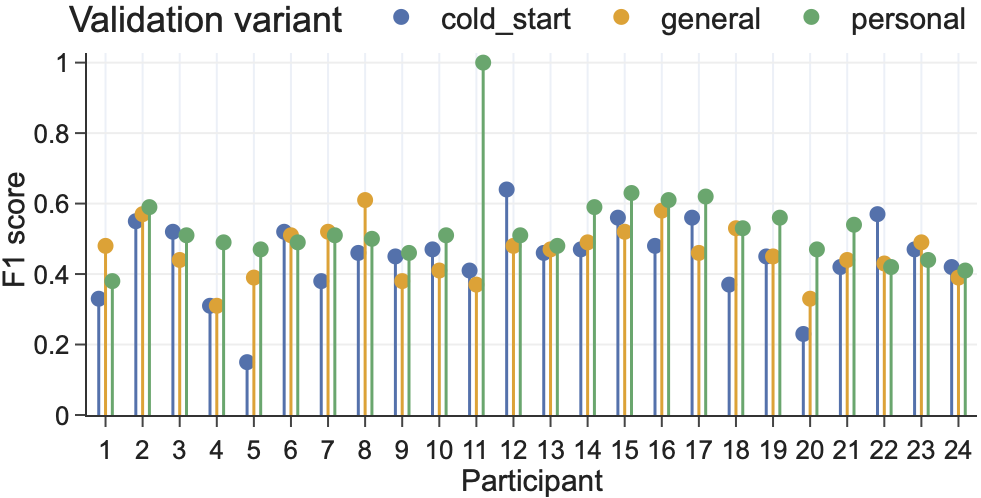}
    \vspace{-3mm} 

  \caption{Per-participant F1-score (macro average) results for finetuned RL-PPG Encoders in different Across-Time validation variants.}  \label{fig:across_time_prcp_resutls}

\end{figure}
    \vspace{-3mm} 
 
While the aggregated metrics demonstrate a clear overarching trend, a participant-level analysis reveals expected inter-subject variability (Fig.~\ref{fig:across_time_prcp_resutls}).  Nevertheless, incorporating any personal data during downstream training yields the highest F1-scores for 19 of 24 subjects. Wilcoxon signed-rank tests across the 24 folds confirm that for the finetuned RL-PPG model the personal variant significantly outperforms all alternatives: cold-start (p=0.004), general (p=0.007), majority (p=0.009), and uniform (p=0.012). Conversely, the general variant shows no significant differences against baselines (all p$\ge$0.36). Ultimately, these findings strongly support our core hypothesis: within the evaluated context, predictive power in highly subjective affective tasks appears to stem primarily from personalized data, rather than general representations or population-level label distributions.

    \vspace{-1mm} 

\section{Discussion and Limitations}
    \vspace{-1mm} 

Consistent with current literature, our LOSO experiments confirm that SSL representations effectively transfer to real-world PPG tasks with objective ground truths. In our physical activity recognition sanity check, the finetuned SSL delivered an almost 5-fold increase in F1-score over uniform baselines in the multiclass classification (Tab. \ref{tab:loso_physical_activity}). This confirms that representations learned from everyday physiological data successfully capture meaningful, universally transferable PPG structures, even in the presence of severe noise and inter-subject variability.

In contrast, subjective intense emotion detection did not benefit from generalized SSL pretraining. Our LOSO evaluation revealed severe cross-subject generalization limitations, underscoring that affective experiences are inherently subjective constructs with profound inter-subject variability. Overall, results for real-life emotion detection with general SSL models remain unsatisfactory, underscoring the need for better, more tailored solutions. As demonstrated by our Across-Time experiments, the leading driver of classification performance is personal data, not a general SSL representation. Subject-specific labeled samples allow the model to bypass population-level noise and capture highly individualized patterns. While tested on just a single dataset, these promising results suggest a shift toward personalized SSL approaches. This hypothesis is well-supported by recent work highlighting personalization as a crucial factor for effective affect recognition from physiological signals \cite{zhao2018personality, han2024systematic}.

Several limitations should be acknowledged. A primary constraint is that we evaluated our approach on a single real-life emotion dataset, a consequence of the lack of fully unconstrained, open-access longitudinal affective datasets with physiological signals. Even within this longitudinal dataset, the labeled data volume becomes relatively small when partitioned for subject-level personalization. Methodologically, relying on fixed 10-second windows may not optimally capture the complex temporal dynamics of affective responses. Furthermore, our unimodal approach utilizes only the PPG signal. Incorporating accelerometry could provide crucial behavioral context for real-world settings, e.g., emotional gesticulation. 

We must also acknowledge the inherent physiological ambiguity of real-world data. PPG changes corresponding to reported intense emotions may be absent or heavily obscured. Actual physiological arousal might be too low to manifest cardiovascularly, or unrecorded contextual factors may override the affective response. Fully resolving this requires capturing an impossibly comprehensive set of variables, representing a fundamental limit of ambulatory sensing. Ultimately, substantial computational costs restricted our analysis to a single SSL backbone, making the investigation of alternative architectures and pretraining strategies a highly promising future direction.

These limitations directly motivate future work, including exploring variable window lengths, evaluating alternative architectures, and investigating pretraining on mixed real-life and laboratory datasets. Furthermore, while developing fully personalized SSL models pretrained exclusively on an individual's unlabeled data is theoretically optimal for subjective tasks, acquiring the necessary longitudinal data remains a severe practical bottleneck. To bridge this gap, group-based personalization approaches~\cite{saganowski2022cold} offer a highly promising solution. By leveraging demographic information or personality traits, these approaches can balance deep personalization with realistic data constraints.
    \vspace{-2mm} 

\section{Conclusions}
    \vspace{-1mm} 

The design of affective AI systems should be guided by ecological validity and intended real-world applications. Our findings indicate that within the evaluated setup, generalized SSL models capture transferable physiological patterns for objective physical tasks, yet they exhibit limitations when applied to the subjective nuances of ambulatory emotion detection. The severe cross-subject generalization gap observed in our subjective tasks underscores that personalization is likely a necessary focus for practical affect recognition in the wild. Although collecting fully unconstrained, real-life physiological datasets is resource-intensive, continued investment from the research community is essential to validate these personalized approaches. Addressing both the data scarcity and the need for individualized modeling architectures will be a crucial next step toward developing ethical, inclusive, and unbiased affective machine learning systems capable of genuinely impacting human mental health and well-being.

\newpage

\section*{Ethical Impact Statement}
    \vspace{-1mm} 

The use of wearable devices and machine learning models for emotion recognition raises important ethical considerations, including data privacy, potential model biases, and the risk of misinterpretation or unfair profiling of individuals. To mitigate these risks, all datasets used in this work were collected in accordance with protocols approved by the Wrocław Tech Ethics Committee (approval no. O-23-08). This ensures that participant consent, data handling, and study procedures met established ethical standards \cite{behnke2022ethical}. Participants were fully informed about the study procedures and their right to withdraw at any time. To balance withdrawal rights with privacy, participant identifiers were retained for a predefined period of two years. After this deadline, all data was completely anonymized, making it impossible to map records back to specific individuals. The compensation for participating in the real-life emotion dataset collection was designed to be fair and aligned with the country's standards. The compensation structure accounted for compliance and wear-time, as detailed in Table \ref{tab:compensation}. The physical activity recognition laboratory dataset was collected during internal smartwatch data quality tests under the same Ethical Committee approval. As these participants were actively involved in the research project, they volunteered without financial compensation. All informed consent procedures were strictly adhered to.

\begin{table}[htpb]
 \vspace{-5mm} 

\caption{Participant compensation structure.}
\centering
\begin{tabular}{llc}
\toprule
\makecell{\textbf{Compensation}\\ \textbf{Component}} & \makecell{\textbf{Compliance} \\ \textbf{Requirement}} & \makecell{\textbf{Amount}\\\textbf{(USD)}} \\
\midrule
\textbf{Base Pay} & Standard participation & \$60 \\
\midrule
\textbf{Emotion Surveys Bonus} & $\ge 60\%$ completion & +\$20 \\
(system-triggered only)        & $\ge 40\%$ completion  & +\$10 \\
\midrule
\textbf{Daily Surveys Bonus}   & $\ge 85\%$ completion    & +\$20 \\
 (morning and evening)         & $\ge 70\%$ completion    & +\$10\\
\midrule
\textbf{Physiology Bonus}      & $\ge 24$ valid days  & +\$20 \\
 $\ge$ 6 hours wear time)      & $\ge 22$ valid days  & +\$10 \\
\midrule
\textbf{Maximum Total Earnings}& Base + All Bonuses                 & \textbf{\$150} \\
\bottomrule
\end{tabular}
\label{tab:compensation}
 \vspace{-3mm} 

\end{table}

In the spirit of transparency and open science, we emphasize our commitment to making the Real-life Emotion Dataset accessible to the broader research community. While the data is not yet publicly available, we are currently finalizing the necessary procedures to ensure the dataset can be safely and openly shared in the near future.

We must acknowledge several limitations regarding diversity, representation, and hardware bias. Firstly, the dataset consists of healthy participants who share a single nationality and language. While our findings emphasize that emotion is profoundly subjective, an individual's personal baseline is undeniably shaped by their broader cultural context. Therefore, when designing future ambulatory affective studies, we strongly encourage international collaborations to collect data across multiple geographic locations using identical, standardized protocols. Building these multi-site, cross-cultural datasets would allow the community to investigate better how different cultural backgrounds influence the dynamics of personalized modeling.

Secondly,  individuals with cardiovascular disease or mental health disorders were not included in the study. Because our affective modeling relies heavily on cardiovascular signals (PPG), pre-existing cardiac conditions or related medications would fundamentally alter a user's physiological baseline. Similarly, clinical mental health disorders can profoundly shift affective reactivity and subjective reporting. Consequently, we strongly emphasize that individuals with such conditions should not use emotion recognition systems trained on healthy populations, as the resulting inferences could be highly misleading or misinterpreted. For clinical populations, affective computing systems must not rely on off-the-shelf generalized models. Instead, they require purpose-built datasets, rigorous testing against clinical ground truths, and development in close collaboration with medical professionals to ensure both efficacy and user safety. 

Thirdly, PPG inherently suffers from optical limitations. We had to exclude participants with a BMI of 30 or higher due to signal degradation. Moreover, PPG sensors can exhibit performance disparities across darker skin tones due to their light-absorption characteristics. To prevent these hardware biases, future affective systems must move beyond single-modality optical sensing. We strongly recommend adopting multimodal inference architectures that pair PPG with physiological sensors unaffected by optical properties, such as electrodermal activity (EDA). Furthermore, alongside further PPG sensor development, leveraging alternative data sources, like behavioral or contextual data, could help ensure that emotion recognition technologies are equitable, resilient, and effective for all body types and demographics.

Our findings underscore the necessity of personal data for subjective tasks. However, we recognize the limitations of this approach. Capturing a participant's full personal context, including their current health, socioeconomic situation, beliefs, etc., is practically impossible. Consequently, affective inference will always lack certain crucial contextual information. Furthermore, subjective self-assessments are susceptible to temporal drift. A participant's perception of events changes over their lifespan, for example, a subject might rate finding \$5 on the street as 100\% positive valence and later rate winning the lottery with the same score, despite likely differences in the absolute impact of the event. Future personalized models should explore time-based alignment, determining how many recent days, months, or years of data should be included to keep the model properly calibrated to the user's current psychophysiological baseline.

Finally, we recognize the dual-use nature of emotion recognition systems, which could be misused for surveillance or to cause harm. We align with the Affective Computing community's commitment to transparency, participant protection, and responsible innovation. By openly sharing both our findings and methodological limitations, we hope to help the community learn from these challenges and develop more resilient technical and ethical safeguards against potential misuse.

\bibliography{bibliography}

\bibliographystyle{IEEEtran}

\end{document}